\documentclass[twoside]{article}

\usepackage[preprint]{Styles/aistats2026}

\usepackage[
  backend=biber,
  style=authoryear,     
  natbib=true,           
  maxcitenames=1,       
  mincitenames=1,      
  maxbibnames=99,       
  uniquelist=false    
]{biblatex}
\usepackage{todonotes} 
\usepackage{graphicx}  
\usepackage{amsmath,amssymb,amsthm}
\usepackage{gensymb} 
\usepackage{booktabs} 
\usepackage{multirow} 
\usepackage{float}    
\usepackage{stfloats} 
\usepackage{hyperref}

\begin{document}

\twocolumn[

\aistatstitle{LatticeVision: Image to Image Networks for Modeling Non-Stationary Spatial Data}

\aistatsauthor{ Antony Sikorski$^{1}$ \And Michael Ivanitskiy$^{1}$ \And  Nathan Lenssen$^{1,2}$} 
\vspace{2pt}
\aistatsauthor{Douglas Nychka$^{1}$ \And Daniel McKenzie$^{1}$}

\aistatsaddress{ Colorado School of Mines$^{1}$ \\  NSF National Center for Atmospheric Research$^{2}$} ]

\begin{abstract}
  In many applications, we wish to fit a parametric statistical model to a small ensemble of spatially distributed random variables (`fields'). However, parameter inference using maximum likelihood estimation (MLE) is computationally prohibitive, especially for large, non-stationary fields. Thus, many recent works train neural networks to estimate parameters given spatial fields as input, sidestepping MLE completely. In this work we focus on a popular class of parametric, spatially autoregressive (SAR) models. We make a simple yet impactful observation; because the SAR parameters can be arranged on a regular grid, both inputs (spatial fields) and outputs (model parameters) can be viewed as images. Using this insight, we demonstrate that image-to-image (I2I) networks enable faster and more accurate parameter estimation for a class of non-stationary SAR models with unprecedented complexity. \looseness=-1
\end{abstract}

\section{INTRODUCTION}
\label{sec:Intro} 

Modeling large, gridded spatial data has become a central challenge in many scientific and industrial applications. This typically involves fitting parametric spatial models to enable prediction, data fusion, and, when data are limited, rapid simulation of additional fields. The bottleneck in this framework is inferring the parameters of the statistical model using maximum likelihood estimation (MLE), which becomes computationally intractable as dataset size increases \parencite{stein2008modeling, sungeostat}. Moreover, spatial data over large domains frequently exhibit \textit{non-stationarity}, meaning key parameters vary over space, further complicating parameter estimation. Many recent works have replaced MLE with neural networks, mapping spatial fields directly to local parameter estimates \parencite{liu2020task,banesh2021fast,zammit2024neural,lenzi2023neural}. Like MLE, these methods are limited to dividing large fields into smaller sections, each assumed to be stationary, and estimating parameters independently for each section. Although faster than MLE, the number of forward passes required scales linearly with the number of sections. Additionally, such local neural estimators struggle to capture long-range, global context.

\begin{figure*}
    \centering
    \includegraphics[width=0.77\textwidth]{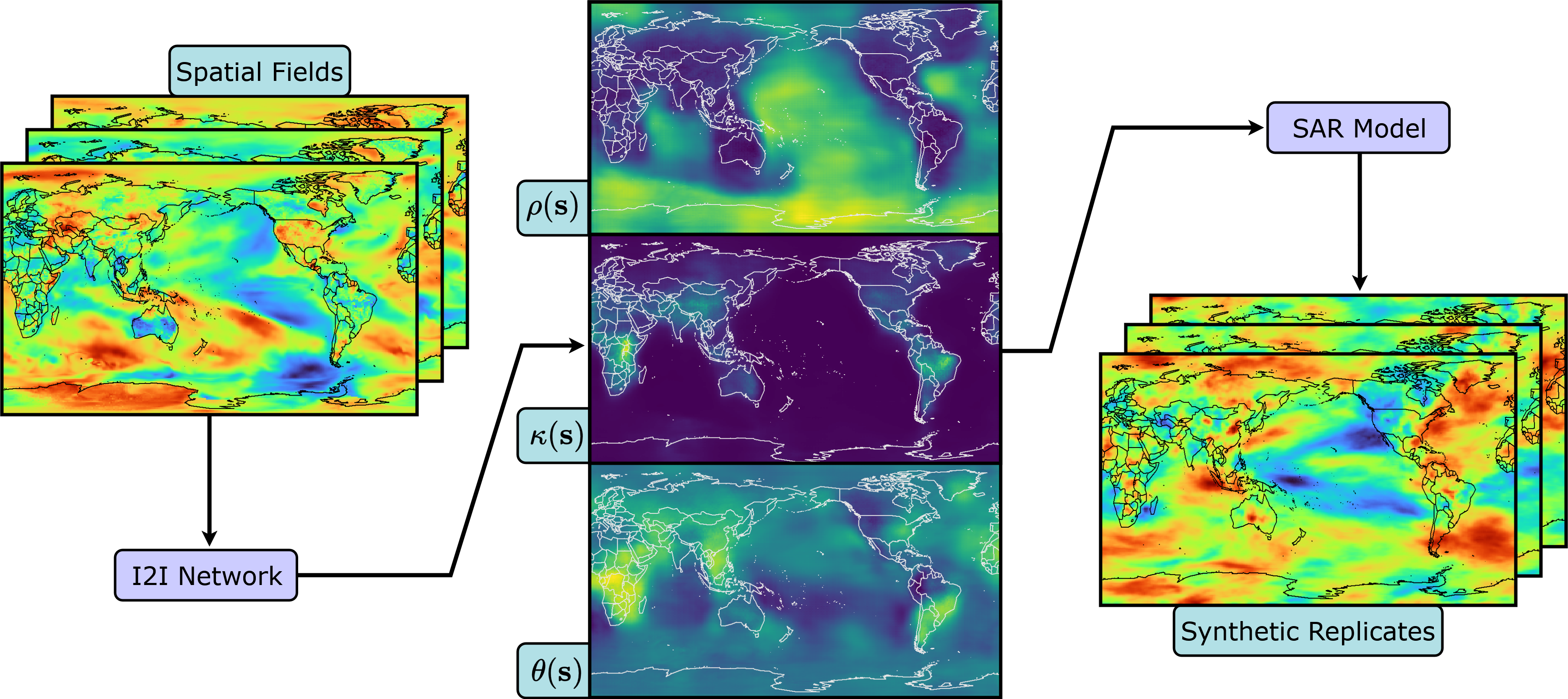}
    \caption{An illustration of the LatticeVision workflow applied to ESM outputs. Spatial fields are fed into an I2I network, which in turn produces estimates of the non-stationary parameter fields. These are encoded into a SAR model from which synthetic replicates are efficiently simulated.}
    \label{fig:flowchart}
\end{figure*}

In this work we introduce LatticeVision, a global estimation and emulation framework for large, non-stationary spatial data. We observe that for a popular group of statistical models known as spatial autoregressive (SAR) models, the parameters themselves are naturally arranged on a grid. {\em Thus, both the spatial fields of interest and their associated parameters can be viewed as images}. Consequently, we adapt image-to-image (I2I) networks---both fully convolutional \parencite{ronneberger2015u} and vision transformer based \parencite{dosovitskiy2020image,chen2021transunet}---to the parameter estimation task. Unlike local neural estimators, I2I networks estimate all model parameters at once, in a single forward pass. Our networks are chosen to evaluate whether incorporating the attention mechanism improves performance over purely convolutional approaches \parencite{liu2022convnet}, which have far fewer parameters. We find that a hybrid approach offers the best performance. \looseness=-1 

A key challenge in training these networks is ensuring they recognize complex non-stationarity patterns encountered in large, geoscientific data. Since we require (field, parameters) pairs for training, the existing corpora of application specific fields \parencite{nguyen2023climax,kaltenborn2023climateset,watson2022climatebench} are inapplicable. Thus we generate our own training data, encoding priors that represent the kind of non-stationary spatial processes expected for geophysical variables. 

In experiments with simulated data, I2I networks outperform local neural estimators in both speed and accuracy. This advantage stems from the fact that I2I networks process the entire field at once, rather than by section \parencite{sainsbury2024neural} or by pixel \parencite{wiens2020modeling}. We also show that I2I networks estimate weakly-identifiable parameters from a small number of replicate spatial fields more reliably than previous, local approaches.   

As an illustrative example, we employ the LatticeVision framework on data from multiple Earth System Models (ESMs). ESM simulations (`runs') model the long-term evolution of Earth's climate, providing decision-makers with critical projections.
 The computational cost of performing more than a handful of ESM runs is extremely prohibitive, limiting ensemble sizes to 3-100 members \parencite{kay2015community,rodgers2021ubiquity}; too few for many applications \parencite{milinski2020large, deser2020insights,schwarzwald2022importance,eyring2024pushing}. Following parameter estimation, we generate realistic ensembles containing thousands of fields in a matter of seconds using the LatticeKrig package \parencite{latticekrig}; a stark contrast to the tens of millions of core-hours required by ESMs. We show that ensembles simulated with parameters estimated by I2I networks better capture spatial relationships---especially long-range anisotropic correlations---than those produced by local methods.

In summary, we make the following contributions:
\begin{itemize}
    \item We propose a \textit{global} framework for efficiently estimating non-stationary parameters with I2I networks, demonstrate that hybrid architectures outperform those that are purely convolutional or transformer-based for this task, and address the limitations of existing \textit{local} approaches. 
    \item We provide a strategy for generating training data that encodes scientifically meaningful priors that future estimators and emulators can use. 
    \item We pair our novel, global estimators with a computationally efficient and flexible SAR model \parencite{latticekrig,LKrigpackage}, and validate this framework on ESM outputs.      
\end{itemize}
All of the accompanying code is available at \href{https://github.com/antonyxsik/LatticeVision}{github.com/antonyxsik/LatticeVision}. \looseness = -1

\section{BACKGROUND: GAUSSIAN PROCESSES AND SAR MODELS}
\label{sec:Background}
\begin{figure*}[h]
    \centering
    \includegraphics[width=0.80\textwidth]{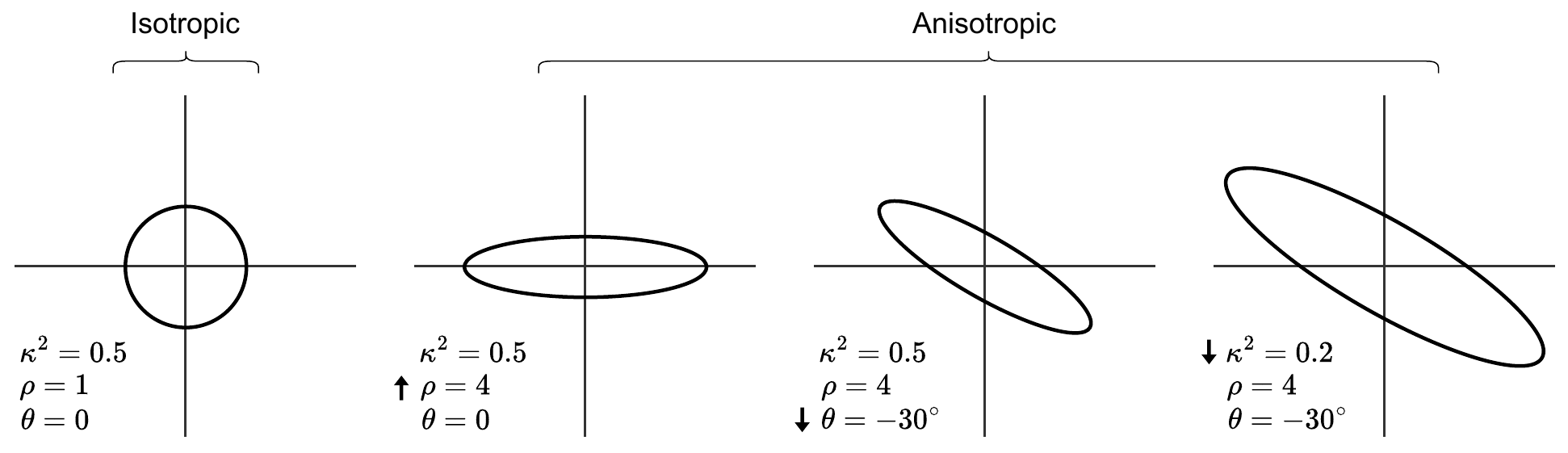}
    \caption{Illustration of the effects of $\kappa^2, \rho$, and $\theta$. The ellipses represent contours of constant correlation, e.g. all locations with correlation $0.5$ with the origin. $\kappa^2$ controls the radii of the ellipse, $\rho$ controls the ratio of the semi-major and semi-minor radii (i.e., the `aspect ratio' of the ellipse), and $\theta$ is the angle the semi-major ellipse makes with the positive $x$-axis.}
    \label{fig:aniso}
\end{figure*}
The past thirty years have seen the development of statistical models for spatial data that are invaluable for spatial prediction and emulation \parencite{heaton2019case, katzfuss2017multi, fuentes2002spectral, guhaniyogi2018meta}. Despite recent advances in generative deep learning for spatial data \parencite{ruhling2023dyffusion, price2023gencast, li2023seeds}, statistical models outperform other approaches when data is limited \parencite{lutjens2025impact}. Moreover, statistical models contain interpretable parameters, which are useful for downstream applications.

We consider statistical models built around Gaussian processes (GPs). With locations $\mathbf{s} \in \mathbb{R}^2$, a mean-zero GP $f(\mathbf{s})$ is fully specified by its covariance function $k(\mathbf{s}, \mathbf{s}') = \mathbb{E}\left[f(\mathbf{s})f(\mathbf{s}')\right]$. The kernel $k$ must be positive definite and this requirement is typically 
enforced by assuming {\it stationarity} and {\it isotropy} \parencite{cressie2015statistics}. That is, $k$ is independent of the location in the domain (stationarity), and depends only on the separation distance (isotropy):
\begin{equation}
 k(\mathbf{s}, \mathbf{s}^{\prime})=\sigma^2 \mathcal{C}( \kappa_m\|\mathbf{s} - \mathbf{s}'\|).
 \label{eq:stat}
 \end{equation} 
Here $\mathcal{C}(0) = 1$, $\sigma^2$ is the variance of the GP, and $\kappa_m > 0$ governs the spatial correlation range. A popular choice for $\mathcal{C}$ is the Matérn family\footnote{In reviewing commonly used covariance kernels, \cite{stein1999interpolation} concludes: `Use the Matérn'.} with an additional smoothness parameter, $\nu>0$. The Matérn class includes several common kernels as special cases (e.g., exponential and Gaussian). The primary obstacle to working with GP-based models is the computational cost (\(\mathcal{O}(n^3)\) operations and \(\mathcal{O}(n^2)\) memory for $n$ spatial locations) of factorizing the covariance matrix $\Sigma \in \mathbb{R}^{n\times n}$ where $\Sigma_{ij} = k(\mathbf{s}_i,\mathbf{s}_j)$ \parencite{sungeostat}.\looseness=-1

\paragraph{The SPDE Method} GPs from the Matérn family have an equivalent  representation in terms of a stochastic partial differential equation (SPDE)
 \parencite{matern2013spatial,whittle1954stationary}:
\begin{equation}
(\kappa^2 - \Delta)^{(\nu+1)/2} f(\mathbf{s}) = \mathcal{W}(\mathbf{s}),
\label{eq:spde}
\end{equation}
where $\kappa^2$ controls the correlation range, and is similar but not identical to  $\kappa_m$ in the Matérn, $\Delta$ is the Laplacian operator, and $\mathcal{W}(\mathbf{s})$ is a white noise Gaussian process with zero mean and variance $\sigma^2$. The so-called SPDE method \parencite{lindgren2011explicit,lindgren2022spde} connects Matérn GPs to Gaussian Markov Random Fields (GMRFs) by demonstrating that discretizing Equation \eqref{eq:spde} on a regular grid yields a GMRF which approximates the Matérn GP well. This approach can be extended\footnote{For simplicity, we shall henceforth focus on the $\nu=1$ case, the so-called Whittle covariance \parencite{whittle1954stationary}} to allow non-stationarity and anisotropy by incorporating a spatially varying dispersion matrix $D(\mathbf{s}) \in \mathbb{R}^{2\times 2}$ and letting $\kappa^2$ vary in space:\looseness=-1
\begin{equation}
(\kappa^2(\mathbf{s}) - \nabla \cdot D(\mathbf{s}) \nabla) f(\mathbf{s}) = \mathcal{W}(\mathbf{s}).
\label{eq:spde_aniso}
\end{equation}
Following \cite{haskard2007anisotropic}, we construct $D(\mathbf{s})$ via its eigendecomposition:
$D(\mathbf{s}) = R(\mathbf{s})^{\top}\Lambda(\mathbf{s})R(\mathbf{s})$ where
\begin{equation}
R(\mathbf{s}) =
\begin{bmatrix}
\cos\theta(\mathbf{s}) & -\sin\theta(\mathbf{s})\\
\sin\theta(\mathbf{s}) &  \cos\theta(\mathbf{s})
\end{bmatrix},\\
\Lambda(\mathbf{s}) =
\begin{bmatrix}
\rho(\mathbf{s}) & 0\\
0 & \frac{1}{\rho(\mathbf{s})}
\end{bmatrix}
\label{eq:rot_diag_mat}
\end{equation}
The generalized Laplacian in \eqref{eq:spde_aniso} can be written as
\begin{equation}
    \nabla \cdot D(\mathbf{s}) \nabla \equiv D_{1,1}(\mathbf{s}) \frac{\partial^2}{\partial x^2}+2 D_{2,1}(\mathbf{s}) \frac{\partial^2}{\partial x \partial y}+D_{2,2}(\mathbf{s}) \frac{\partial^2}{\partial y^2}.
    \label{eq:H_laplace}
\end{equation}
\paragraph{Interpreting the Parameter Fields} By specifying the ``parameter fields'' $\kappa^2(\mathbf{s}), \theta(\mathbf{s})$, and $\rho(\mathbf{s})$ we obtain a rich class of nonstationary GPs. Because we are defining this model in terms of \eqref{eq:spde_aniso}, the problem of explicitly specifying an analytical form for the covariance that is positive definite is avoided.  Moreover, these parameter fields are interpretable and can yield physical insights into the spatial dependence of the field. Specifically, $\kappa^2$ controls the overall range of correlation (larger $\kappa^2$ means more localized  dependence), $\rho$ controls the degree of anisotropy ($\rho=1$ corresponds to isotropy), and $\theta$ controls the direction of anisotropy; see Figure \ref{fig:aniso}.

\paragraph{Discretizing the SPDE} To obtain a computable model from \eqref{eq:spde_aniso}, one approximates this SPDE using either a finite element or finite difference method. Following \cite{wiens2020modeling}\footnote{We correct a minor error in their derivation. Our derivation can be found in Appendix \ref{subsec:aniso_sar_appendix}} we use the finite difference method on a regular grid, yielding the stencil:
\begin{equation}
    \begin{array}{c|c|c}
    \frac{D_{1,2}(\mathbf{s})}{2} & -D_{2,2}(\mathbf{s}) & \frac{-D_{1,2}(\mathbf{s})}{2} \\
    \hline -D_{1,1}(\mathbf{s}) & \kappa^2(\mathbf{s}) + 2 D_{1,1}(\mathbf{s}) + 2 D_{2,2}(\mathbf{s}) & -D_{1,1}(\mathbf{s}) \\
    \hline \frac{-D_{1,2}(\mathbf{s})}{2} & -D_{2,2}(\mathbf{s}) & \frac{D_{1,2}(\mathbf{s})}{2}.
    \end{array}
\label{eq:stencil_better}
\end{equation}
Let $\mathbf{y} \in \mathbb{R}^n$ denote a discretized solution to \eqref{eq:spde_aniso} on this grid, with $y_{i,j}$ denoting the value at grid location $(i,j)$. Then $\mathbf{y}$ is the solution to $B\mathbf{y} = \mathbf{e}$ where $\mathbf{e}\sim \mathcal{N}(\mathbf{0},I)$ is a sample from the standard multivariate normal distribution and $B \in \mathbb{R}^{n\times n}$ is the spatial autoregressive (SAR) matrix associated to the stencil \eqref{eq:stencil_better}. \cite{lindgren2011explicit} show that $\mathbf{y}$ approximates a sample from the Matérn GP associated to \eqref{eq:spde_aniso}. As $B$ is sparse and structured ({\em i.e.}, it is a banded matrix) this linear system may be solved at a computational cost $\mathcal{O}(n^{3/2})$, thus sidestepping the bottleneck associated with working with the GP directly. This formulation results in a GMRF with a particular SAR structure, where, by linear statistics, the precision matrix is $Q = B^{\top}B$.

\section{NON-STATIONARY DATA GENERATION} 
\label{sec:DataGen}
\paragraph{I2I Data} 
In order for the I2I networks to be successful, the training data needs to encode priors appropriate for geoscientific applications. Previous work \parencite{wiens2020modeling} has shown that coastlines, long-range East-West correlations due to jet streams, and oceanic circulation yield parameter fields that are important yet challenging to detect. So, we construct a pipeline for generating synthetic fields exhibiting these key phenomena. First, we construct spatially varying parameter fields $\kappa^2(\mathbf{s}), \rho(\mathbf{s}), \theta(\mathbf{s})$ and then use them to produce an $M$-replicate ensemble of synthetic fields $Y = \{\mathbf{y}^{(m)}\}^{M}_{m=1}$, where each $\mathbf{y}^{(m)}\in\mathbb{R}^{H\times W}$ is generated using the same parameter fields. These parameter fields are concatenated along the channel dimension to form a three channel, ground truth ``image'' $\Phi \in \mathbb{R}^{3\times H \times W}$, where $\Phi(\mathbf{s}) = [\kappa^2(\mathbf{s}), \rho(\mathbf{s}), \theta(\mathbf{s})]^T$. The replicates of the synthetic fields are also concatenated along the channel dimension to form the input image to our networks $Y \in \mathbb{R}^{M\times H \times W}$. Thus, we have input-output pairs $(Y,\Phi)$ in our data. 

Each time we construct a single parameter field, we first sample one of eight spatial patterns $p^{(i)}(\mathbf{s}; \boldsymbol{\Omega}^{(i)})$, $ i = 1,2,\dots,8$. The patterns are simple spatial functions that dictate how a parameter will vary across the domain, and are designed to be caricatures of real geophysical variability. We hypothesize that these patterns will serve as ``building blocks'', enabling our networks to generalize to the kinds of parameter changes present in geophysical settings. Each pattern is defined by its hyperparameters $\boldsymbol{\Omega}^{(i)}$, which are randomly sampled from a series of prior uniform distributions. For example, when a ``Coastline'' pattern is selected, values that dictate the position and variation of the ``Coastline'' are chosen as well. The relative frequency and qualitative descriptions of these patterns are illustrated in Figure \ref{fig:configs}, with detailed functional forms and hyperparameter priors provided in Appendix \ref{sec:data_generation_appendix}.
\begin{figure}[h]
    \centering
    \includegraphics[width=0.44\textwidth]{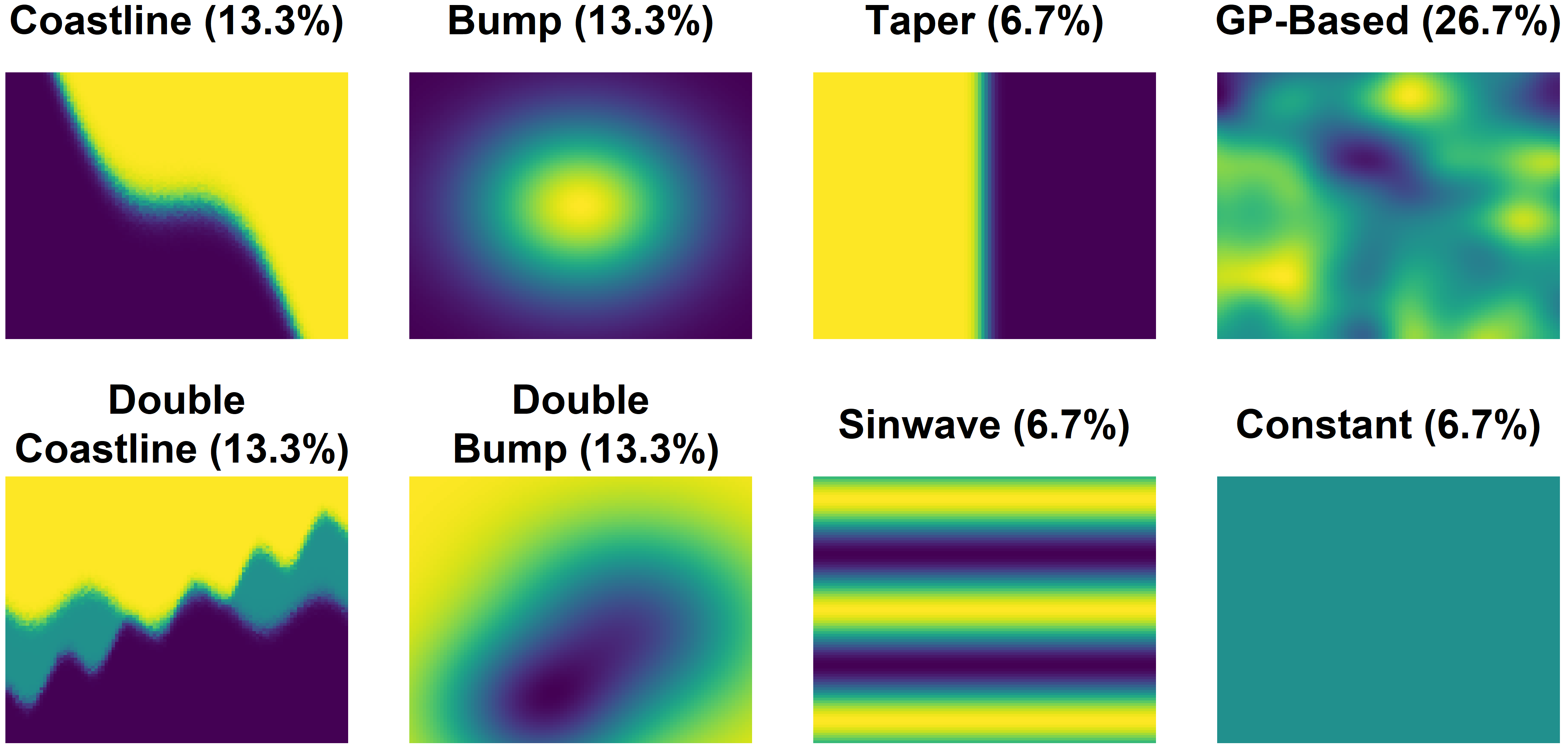}
    \caption{Spatial patterns and their frequencies.}
    \label{fig:configs}
\end{figure}
Once a pattern $p^{(i)}$ and its hyperparameters $\boldsymbol{\Omega}^{(i)}$ have been chosen, we sample values from the prior distribution of the specific parameter $(\kappa^2,\rho,\text{or } \theta)$ for which we are constructing a field. Each pattern (except ``Constant'') requires sampling two values which dictate the maximum and minimum value across the resulting parameter field. We set uniform priors on the anisotropy parameters $\rho \sim \mathcal{U}(1, 7)$ and $\theta \sim \mathcal{U}\left(-\frac{\pi}{2}, \frac{\pi}{2}\right)$, and set a mixed prior of $\kappa^2 \sim 0.6\log\mathcal{U}(10^{-4}, 2) + 0.4\mathcal{U}(10^{-4},2)$ for the correlation range. These choices ensure we capture a broad range of spatial relationships, ranging from isotropy ($\rho = 1$) to very elongated ellipses ($\rho \gg 1$) in any direction. 

We follow this process each time to create a parameter field for $\kappa^2(\mathbf{s}), \rho(\mathbf{s})$, and $ \theta(\mathbf{s})$, and then encode all parameters into the SAR matrix $B$, as described in Section \ref{sec:Background}. Drawing white noise $\mathbf{e}\sim \mathcal{N}(0,I)$ and solving $B\,\mathbf{y} = \mathbf{e}$ yields a random field $\mathbf{y}$ with the desired spatial covariance structure. Repeating this for $M$ independent draws produces a small ensemble $Y = \{\mathbf{y}^{(m)}\}^{M}_{m=1}$ of different synthetic fields with identical covariance structures. 
\paragraph{CNN Data}
For the local estimation setting, we assume local stationarity: within a small spatial window, all grid cells are governed by the same  parameters. This makes data generation for the local CNNs comparatively straightforward. Rather than constructing parameter fields, we independently sample a single value for each of $(\kappa^2, \rho, \theta)$ from the prior distributions defined above. These parameters are then encoded into a SAR matrix $B$, resulting in stationary synthetic fields. We generate smaller fields than for the I2I networks, with each sample consisting of $M$ replicates $\mathcal{Y} \in \mathbb{R}^{M \times h \times w}$, where $h \ll H$ and $w \ll W$. The associated ground truth is a parameter vector $\phi = [\kappa^2, \rho, \theta]^T \in \mathbb{R}^3$. Thus, our input-output pairs are $(\mathcal{Y}, \phi)$.\looseness=-1

\section{PARAMETER ESTIMATION NETWORKS}
\label{sec:EstimationNetworks}
We adapt three I2I networks for parameter estimation: a fully convolutional U-Net, a modified ViT, and a hybrid network inspired by the TransUNet architecture. A range of CNNs representative of the local neural estimator literature serve as our baselines. 
\paragraph{UNet} We use a standard U-Net \parencite{ronneberger2015u} with a symmetric encoder-bottleneck-decoder structure and skip connections that propagate spatial information across resolutions. Our implementation replaces ReLU with GELU activations \parencite{gelu}, employs group normalization \parencite{groupnorm}, and omits dropout \parencite{srivastava2014dropout}. The number of channels in the bottleneck matches the transformer embedding dimension used in the transformer-based I2I networks.  

\paragraph{ViT} The original ViT \parencite{dosovitskiy2020image} was designed for image classification, so we remove classification-specific components such as the class token and final MLP head, and use a learnable linear layer to project the transformer output back to the original image resolution. The original 1D positional embeddings are replaced with a range of 2D alternatives; see Section~\ref{sec:SimData}.

\paragraph{STUN} Our hybrid architecture, a spatial TransUNet (STUN), is based upon a network originally designed for image segmentation \parencite{chen2021transunet}. It blends the two I2I networks discussed above, originally combining a pretrained convolutional encoder with a vision transformer and a shallow decoder. We retain the overall structure, but replace the encoder and decoder with the symmetric U-Net components used in our fully convolutional network.

\paragraph{CNN} Our local neural estimators follow the standard design of convolutional layers and a max pooling layer followed by an MLP. We use three networks with varying receptive window sizes that are representative of the local estimation literature \parencite{banesh2021fast, lenzi2023neural, gerber2021fast, sainsbury2024likelihood, wiens2020modeling}. The key difference is our networks have more parameters (1-2.5M), as compared to typical local estimators (50-700k).

\subsection{Implementation and Training Details}

All networks are trained with a batch size of $b=64$ and varying numbers of replicate input fields $M \in \{1, 5, 15, 30\}$. Unless otherwise noted, all reported network sizes and metrics correspond to $M=30$ replicates. For the I2I networks, each training example consists of $M$ input fields of shape $[b, M, H, W] = [64, M, 192, 288]$, with corresponding output parameter fields $[64, 3, 192, 288]$. For the local CNN estimators, we vary the receptive field $h = w = \{9,17,25\}$, with inputs of shape $[64, M, h, w]$ and scalar outputs $[64, 3]$. We append the window size to the name of the CNN, thus $h=w=25$ is denoted CNN25. To encourage permutation invariance, we randomly shuffle the replicates across the channel dimension each training step.\looseness=-1

Certain data augmentation techniques do not preserve the statistical structure of the fields. For example, image rotation does not preserve the angle of anisotropy $\theta$. Thus, we limit augmentation to spatial translation and random field negation. We generate datasets consisting of 8000 (I2I) and 80000 (CNN) samples with 30 replicates, and employ a 90/8/2 (train/validation/test) split, resulting in a test set of 160 samples for the I2I dataset, which both local and global methods are evaluated on. More details on implementation, storage, and software can be found in Appendix \ref{sec:implementation_appendix}.

\section{SIMULATED DATA EXPERIMENTS}
\label{sec:SimData}

\begin{figure}
    \centering
    \includegraphics[width=0.46\textwidth]{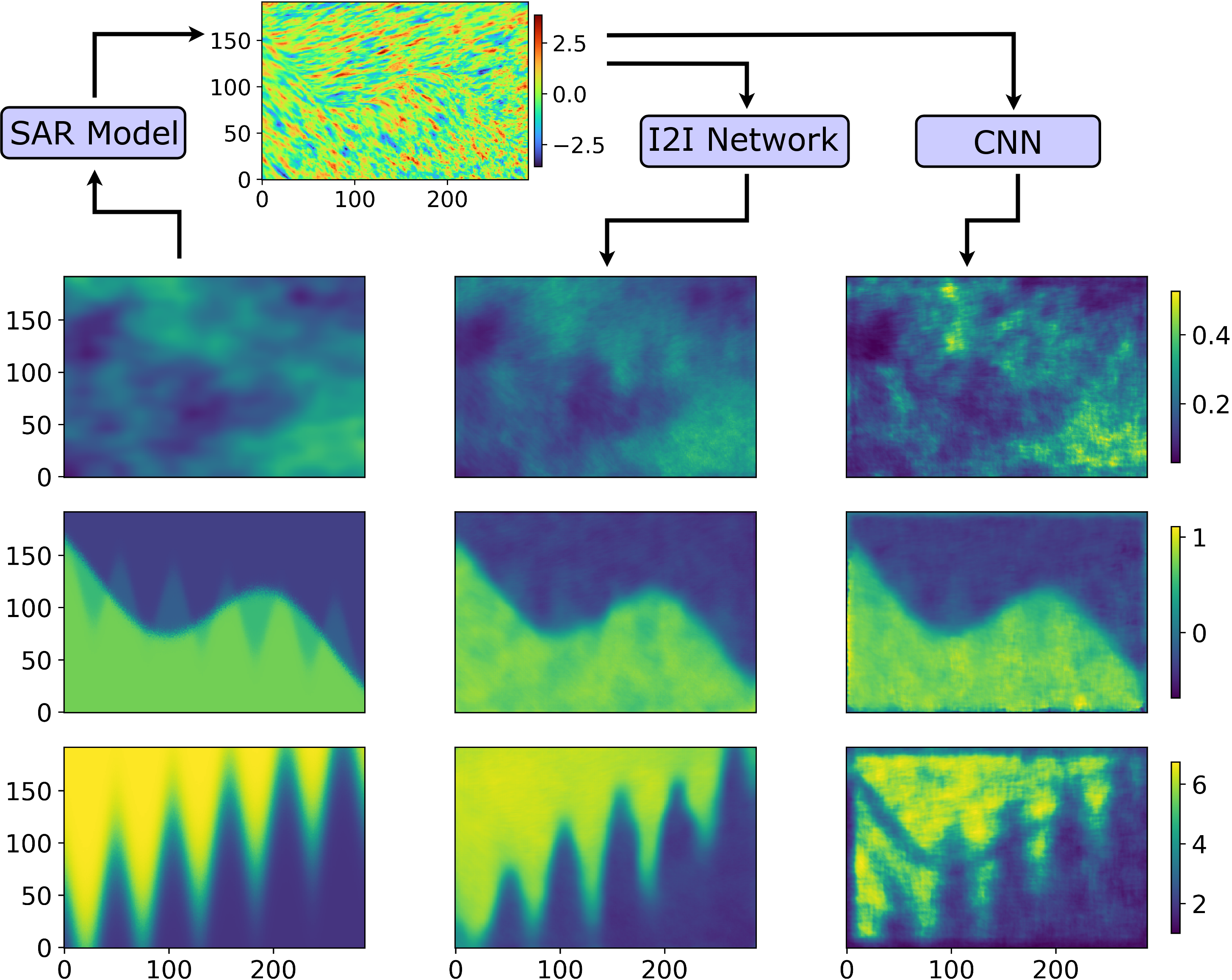}
    \caption{True parameters $\Phi$ (\textbf{left}) are encoded into the LatticeKrig SAR model to simulate a testing sample $Y$, of which one replicate 
    $\mathbf{y}^{(0)}$ is displayed (\textbf{top}). $Y$ is used as an input to STUN and a sliding window, local estimation strategy using CNN25, resulting in $\hat{\Phi}_{\text{STUN}}$ (\textbf{middle}), and $\hat{\Phi}_{\text{CNN25}}$ (\textbf{right}).} 
    \label{fig:sim_estimates}
\end{figure}

We generate a test set of $160$ samples following the procedure in Section \ref{sec:DataGen}. Each sample is a small ensemble \(Y\in\mathbb{R}^{M\times192\times288}\) of \(M\) replicates and an associated parameter image \(\Phi\in\mathbb{R}^{3\times192\times288}\). For the I2I networks (UNet, ViT, STUN) we simply forward–propagate \(Y\) to obtain \(\hat{\Phi}_{\text{I2I}}\). For the local CNN baselines, we employ a pixel-by-pixel approach: we translate the CNN window across the field with a stride of 1, assigning the prediction to the central pixel of the window to build \(\hat{\Phi}_{\text{CNN}}\). We use reflection padding---improving upon prior works which use zero padding---thus reducing edge-artifacts. Figure \ref{fig:sim_estimates} contrasts an example $\Phi$ with estimates from the best performing global (STUN) and local (CNN25) networks. MLE is not attempted; local likelihood evaluation for one field would require hundreds of hours, exceeding the total runtime of our approach by orders of magnitude \parencite{wiens2020modeling}. 

\paragraph{Positional Embeddings}
The spatial nature of our problem leads us to study the effects of positional embeddings. We evaluate four embeddings for the transformer: none, 2-D sinusoidal \parencite{parmar2018image}, learned 2-D \parencite{dosovitskiy2020image}, and rotary (RoPE) \parencite{su2024roformer}. RoPE yields the best performance across the majority of metrics for both ViT and STUN, and is adopted henceforth (see Appendix Table \ref{tab:pos_embed_metrics}).

\paragraph{Number of Replicates}
We observe the effect on parameter estimation performance with varying numbers of replicates $M = \{1,5,15,30\}$. Table \ref{tab:rmse_results} compares results for the extremes of this range. We find that I2I networks are more resilient to a low number of replicates, with STUN and UNet showing almost no difference in prediction RMSE. Full results are in Appendix Tables \ref{tab:appendix_i2i}, \ref{tab:appendix_cnn}.

	\begin{table*}
	\caption{Root mean square error (RMSE) for parameter estimation on a simulated test set using 1 and 30 replicates (reps). ``Size'' is the net parameter count (in millions). ``Train'' is the wall-clock time for training until early stopping, and ``Eval'' is the average time (out of 5) to process the $160$ sample test dataset. Arrows indicate desirable direction, and bold values indicate best performance.}
	\label{tab:rmse_results}
	\centering
	\begin{tabular}{lccc ccc ccc}
	\toprule
	\multirow{2}{*}{Net}&\multicolumn{3}{c}{30 Rep RMSE $\downarrow$}&\multicolumn{3}{c}{1 Rep RMSE $\downarrow$}&\multirow{2}{*}{\shortstack{Size\\(M)$\downarrow$}}&\multirow{2}{*}{\shortstack{Train\\(min)$\downarrow$}}&\multirow{2}{*}{\shortstack{Eval\\(sec)$\downarrow$}}\\
	\cmidrule(lr){2-4}\cmidrule(lr){5-7}
	&$\kappa^2$&$\rho$&$\theta$&$\kappa^2$&$\rho$&$\theta$&&&\\
	\midrule
    CNN9&0.963&0.937&0.316&1.01&1.39&0.535&\textbf{1.3}&\textbf{8}&71.3\\
    CNN17&0.765&0.994&0.293&0.766&1.27&0.443&1.9&19&163.0\\
    CNN25&0.743&1.03&0.272&0.806&1.28&0.418&2.6&35&341.6\\
    ViT&0.374&0.625&0.204&0.471&0.771&0.237&92&161&\textbf{0.30}\\
    UNet&0.201&0.308&\textbf{0.087}&0.195&0.354&0.111&25&144&0.33\\
	STUN&\textbf{0.189}&\textbf{0.302}&0.091&\textbf{0.189}&\textbf{0.351}&\textbf{0.097}&105&178&0.38\\
	\bottomrule
	\end{tabular}
	\end{table*}

Our experiments highlight three consistent trends. (i) Global I2I networks pose a significant improvement over local CNNs: STUN and UNet often display 4-5x lower RMSE and are almost unaffected by shrinking the ensemble size, whereas CNNs are noticeably sensitive. (ii) Adding attention helps only marginally in this setting, and attention on its own lags behind approaches that include multiscale convolutions. STUN edges out UNet but at 4x the parameter count. ViT lags behind the other I2I networks, perhaps as it requires more training data \parencite{dosovitskiy2020image}. (iii) Global I2I networks take longer to train due to a higher parameter count, yet are \textit{amortized} at inference time. In order to achieve the results of a single forward pass through an I2I network, the local estimators must perform $H\times W= 55,296$ forward passes. Consequently, I2I networks perform inference 100--1000 times faster than local neural estimators.\looseness=-1 

\section{CLIMATE APPLICATION}
\label{sec:clim_app}
We evaluate our framework by estimating parameters from climate model outputs, using these parameters to generate large, synthetic ensembles, and then comparing the quality of the I2I-based and local CNN-based emulators. We consider ensembles of surface temperature sensitivity fields from three climate models with differing numbers of replicates and resolutions: MPI-ESM (50 members, $192\times96$ resolution, \cite{olonscheck2023new}), CESM1 (30, $288\times 192$, \cite{kay2015community}), and EC-Earth3 (72, $512 \times 256$, \cite{CMIP6}). While diffusion networks have recently shown promise in adjacent emulation applications \parencite{ruhling2024probablistic,bassetti2023diffesm}, only having 30-72 fields per model renders the training of generative models impractical in this setting.\looseness=-1

The fields represent the local changes in temperature given an increase of $1^{\circ}$C in global temperature. Due to the chaotic nature of the ocean-atmosphere system, we can treat each field within an ensemble as an independent replicate sampled from the model’s ``true'' climate sensitivity. The data is preprocessed into standardized temperature sensitivity anomalies where each pixel has zero mean and unit variance. We then perform parameter estimation with all global I2I and local CNN estimators on 30 randomly selected fields from each ensemble. Regardless of the estimation method used, the LatticeKrig SAR model simulates $1000$ fields in less than one minute on a single laptop CPU, a stark contrast to the tens of millions of core hours required to generate the original ensembles.  

\begin{figure*}[t]
  \centering
  \includegraphics[width=0.81\textwidth]{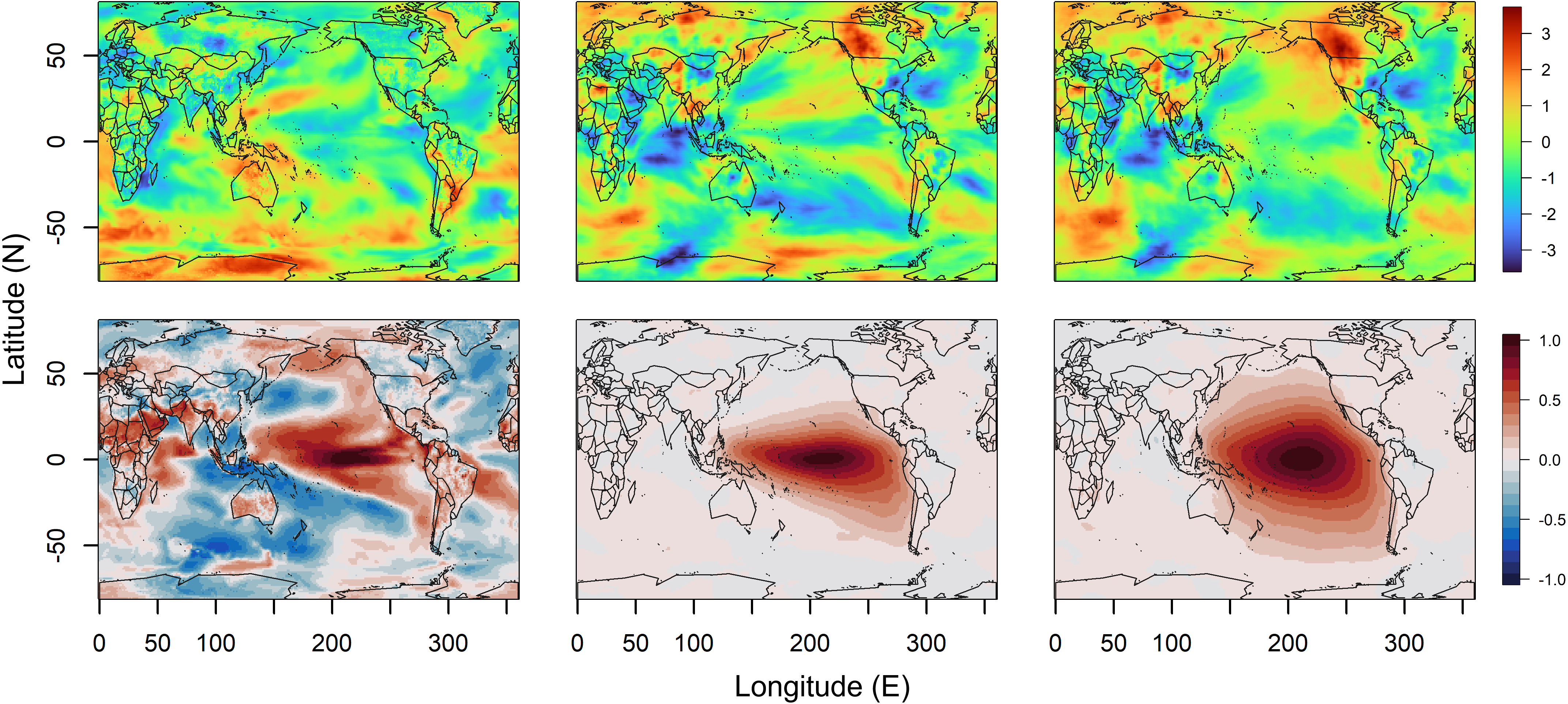}
  \caption{ \textbf{Top:} Standardized temperature fields drawn from the CESM1 ensemble (\textbf{left}), the STUN-based emulator (\textbf{middle}), and the CNN25-based emulator (\textbf{right}). \textbf{Bottom:} Correlations with a chosen location in the Niño 3.4 region at ($212^{\circ}$E, $1^{\circ}$N) for the same three ensembles. The STUN-based emulator better preserves spatial relationships, including the zonal correlation structure along the equator and the meridional oceanic correlation range.}
  \label{fig:clim_covs_fields}
\end{figure*}

For each climate model, we perform the following experiment: We use the I2I-based and CNN-based emulators to generate 1000 fields each, and evaluate how well these synthetic ensembles preserve the spatial relationships present in the climate model outputs. Absolute prediction error is not meaningful here as the ``truth'' is itself a Monte-Carlo sample of internal climate variability. Thus, we compare the second-order structure through a covariance analysis. Rather than computing the entire correlation matrices of the true and simulated fields, which can get as large as  $131,072\times131,072$ (EC-Earth3), we compare a representative sample of $50$ rows.

First, we randomly select 50 anchor locations. We then calculate each anchor's Pearson correlation with every other location for the original fields, and those from the I2I-based and CNN-based emulators. The average RMSE between the 50 true rows and their simulated counterparts is then calculated. {\bf We find that I2I-based emulators consistently outperform CNN-based emulators in capturing spatial relationships}, exhibiting significantly lower RMSEs in eight of nine cases (paired t-tests; Appendix Table \ref{tab:t_tests_clim}). 

As a representative qualitative comparison, we visualize empirical correlations in CESM1 with a chosen point in the Niño 3.4 region—a region of the tropical Pacific most correlated with the El Niño–Southern Oscillation phenomenon \parencite{barnston1997documentation}—for the STUN-based and CNN25-based emulators (Figure \ref{fig:clim_covs_fields}). The STUN-based emulator better preserves the expected zonal east-west correlation structure along the equator with a more realistic correlation range. The CNN-based emulator systematically over-smooths and inflates the oceanic correlation range, particularly in the meridional north-south direction. The original ensemble correlation field is noisier as the CESM1 ensemble contains only 30 members; whereas our synthetic ensembles contain 1,000. These patterns remain consistent across climate models and network pairs. In sum, the I2I-based methods outperform the local CNN-based methods both quantitatively and qualitatively in representing the underlying correlation structure of climate sensitivity fields.

\section{RELATED WORKS}

\paragraph{Neural Parameter Estimation} In situations where the likelihood function is intractable, but simulation from the model is feasible, neural networks have emerged as a powerful alternative to MLE \parencite{liu2020task, zammit2024neural}. Neural parameter estimation combines ideas from simulation-based inference \parencite{cranmer2020frontier,sisson2018handbook} and learning to optimize (L2O) \parencite{chen2022learning, yin2022learning} by training networks to identify maxima of the Bayes risk \parencite{sainsbury2024likelihood}. The training cost is then amortized by repeated use. Until now, these approaches have been limited to local estimation \parencite{banesh2021fast, lenzi2023neural, sainsbury2024likelihood, rai2024fast, walchessen2024neural, gerber2021fast, rai2025extmodeling}, typically using multi-layer perceptrons (MLPs) and CNNs. This line of research is the main inspiration for this work, which advances the field with simultaneous, non-stationary parameter inference and the use of I2I networks. 

\paragraph{Climate Model Applications} 
While climate model emulation is not the sole application of our method, it provides an illustrative example. Physics-based climate models require $10^7$--$10^8$ core hours, necessitating data-driven emulation \parencite{eyring2024pushing}. Despite recent advances in deterministic \parencite{pathak2022fourcastnet,lam2023learning,nguyen2024scaling,bi2023accurate,nathaniel2024chaosbench,keisler2022forecasting, chen2023fengwu} and ensemble-based probabilistic \parencite{kochkov2024neural,price2023gencast,li2023seeds,shi2024codicast} weather forecasting, comparatively fewer methods focus explicitly on long-term climate projections \parencite{lai2024machine, kashinath2021physics}. Our approach draws ideas from statistical methods, which explicitly link parameter estimation with emulation \parencite{castruccio2014statistical, song2024efficient, nychka2018modeling, wiens2020modeling, chakraborty2025learning}, and machine learning (ML) methods, both deterministic \parencite{nguyen2023climax, watt2023ace, chapman2025camulator} and probabilistic \parencite{ruhling2024probablistic, ruhling2023dyffusion} which emulate directly from initial forcings or prior timesteps. Specifically, we combine the straightforward uncertainty quantification and interpretable parameters from the statistical model with a deterministic ML approach for efficient parameter estimation. Purely ML-based climate emulators typically require an extensive corpus of training data \parencite{watson2022climatebench, yu2023climsim} and can exhibit limited generalization capabilities beyond their training distributions \parencite{kaltenborn2023climateset}. We sidestep this by generating synthetic training data that mimics non-stationarity in geophysical settings. Our framework does not serve as a replacement for ESMs, but rather a complementary method for augmenting ensemble sizes, and reducing the number of ``ground truth'' ESM runs that must be computed.\looseness=-1

\section{LIMITATIONS AND EXTENSIONS}
\label{sec:limit_extend}
The limitations of this work fall into two categories: those of the I2I networks, and those of the statistical model.

\paragraph{Estimation Networks} Our I2I networks assume complete, regularly gridded data. Replicate count is also fixed at training time, with only a soft constraint to enforce permutation invariance. A hard constraint via aggregation \parencite{deepsets} might resolve this, and warrants further exploration. Transformer-based networks would likely benefit from larger datasets or advanced training techniques such as student-teacher distillation  \parencite{touvron2021training} and a larger dataset. Future work could extend the data generation pipeline to accommodate variable training dataset dimensions and explore scalable attention mechanisms for larger spatial domains \parencite{cao2022swin}. Our network choices establish a baseline, but there exist many additional architectures for future work to explore \parencite{liu2022swin, rao2022hornet, kim2022instaformer, wang2021pyramid, bao2023channel}.\looseness=-1

\paragraph{Statistical Model} 
We adopt a Gaussian process framework via SAR approximation, which enforces monotonic decay of the covariance and cannot capture teleconnections or nonlinear dynamics such as eddies. Our current formulation omits explicit modeling of an additional white noise process, and could benefit from increased smoothing in areas with long-range correlation structures. Extensions to this work could explore multi-resolution structures \parencite{katzfuss2017multi, frkv2, latticekrig}, estimate a spatially varying noise term, and approximate nonlocal dependence patterns that arise in physical systems.
 
While our framework enables efficient parameter estimation and simulation of plausible ensembles, it ultimately inherits the assumptions and constraints of the underlying statistical model. Applications across a broad range of fields such as  epidemiology, hydrology, or materials science are feasible, although they may require tailoring the data generation strategy and retraining.

\section{CONCLUSION}
We introduce LatticeVision, a global, image-to-image (I2I) framework for the estimation and emulation of non-stationary spatial processes. By representing both the spatial fields and parameters as images, we use I2I networks to estimate all parameters simultaneously, in a single forward pass. We also develop a novel pipeline for generating non-stationary training data. We show that I2I networks demonstrate improvements in accuracy, robustness with few replicates, estimation speed, and ability to capture long-range, anisotropic correlations, as compared to local approaches. We pair I2I parameter estimators with the LatticeKrig SAR model, enabling fast simulation of large ensembles for non-stationary spatial data.

\section{ACKNOWLEDGEMENTS}
We would like to thank Ryan Peterson, Ryker Fish, Sweta Rai, Brandon Knutson, and Samy Wu Fung for their insights and helpful discussions. This material is based upon work supported by the National Science Foundation Graduate Research Fellowship Program under Grant No. DGE-2137099. Any opinions, findings, and conclusions or recommendations expressed in this material are those of the authors and do not necessarily reflect the views of the National Science Foundation. Nathan Lenssen is partially funded through NCAR which is sponsored by the National Science Foundation under Cooperative Agreement 1852977.

\printbibliography

\clearpage
\newpage

\appendix
\thispagestyle{empty}

\onecolumn
\aistatstitle{Appendix}

\section{ANISOTROPIC SAR DERIVATION}
\label{sec:SAR_appendix}

\subsection{The Matérn Covariance}

Under the Matérn family, Equation \eqref{eq:stat} takes the form:

\begin{equation}
k(\mathbf{s},\mathbf{s}^{\prime}) = \sigma^2 \frac{2^{1-\nu}}{{\Gamma}(\nu)} (\kappa_m \|\mathbf{s} - \mathbf{s}'\|)^\nu \mathcal{K}_\nu(\kappa_m \|\mathbf{s} - \mathbf{s}'\|),
\label{eq:matern}
\end{equation}

where $\Gamma(\cdot)$ is the gamma function, and $\mathcal{K}_{\nu}(\cdot)$ is the modified Bessel function of the second kind.

\subsection{Isotropic SAR}

In two dimensions ($\mathbf{s} \in \mathbb{R}^2$), with $\nu = 1$, Equation \eqref{eq:spde} can be written as 

\begin{equation}
(\kappa^2 - \Delta) f(\mathbf{s}) = \mathcal{W}(\mathbf{s}),
\label{eq:spde_appendix}
\end{equation}

where $\Delta=\partial^{2}/\partial x^{2}+\partial^{2}/\partial y^{2}$ is the Laplacian operator. Let the domain be covered by a square lattice with unit spacing.
We denote $f_{i,j}\equiv f(\mathbf{s}_{i,j})$ at grid point
$\mathbf{s}_{i,j}=(i,j)$, $i,j\in\{1,\dots,N\}$.

Using second order central-difference approximations for the Laplacian yields 
\begin{equation}
    -\Delta f_{i,j}\;\approx\;
\bigl[\,4f_{i,j}-f_{i+1,j}-f_{i-1,j}-f_{i,j+1}-f_{i,j-1}\bigr].
\label{eq:iso_approx}
\end{equation}

Substituting \eqref{eq:iso_approx} into \eqref{eq:spde_appendix} results in 
\begin{equation}
    (\kappa^{2}+4)\,f_{i,j}
-\bigl(f_{i+1,j}+f_{i-1,j}+f_{i,j+1}+f_{i,j-1}\bigr)
\;=\;
e_{i,j},
\label{eq:iso-discrete}
\end{equation}
where $e_{i,j}$ denotes the discrete version of the noise \(\mathcal{W}\) at location \((i,j)\). Equation \eqref{eq:iso-discrete} can be visualized in lattice notation using the following stencil:

\begin{equation}
    \begin{array}{c|c|c}
    0 & -1 & 0 \\
    \hline
    -1 & \kappa^2 +4& -1 \\
    \hline
    0 & -1 & 0
    \end{array}
\label{eq:isolattice_appendix}
\end{equation}

illustrating the isotropic relationship between \(f_{i,j}\) and its neighboring locations. 

\subsection{Anisotropic Extension}
\label{subsec:aniso_sar_appendix}

To incorporate geometric anisotropy we replace the Laplacian in
\eqref{eq:spde_appendix} by the \emph{generalised} Laplacian  
\(\nabla\cdot D\nabla\), where the positive–definite dispersion matrix  
\(D\in\mathbb{R}^{2\times2}\) is

\begin{equation}
D \;=\; R^{\top}\Lambda R, \qquad
R=\begin{bmatrix}\cos\theta&-\sin\theta\\[2pt]\sin\theta&\cos\theta\end{bmatrix},
\quad
\Lambda=\begin{bmatrix}\rho&0\\[2pt]0&\frac{1}{\rho}\end{bmatrix}.
\end{equation}

Here, $R$ is a rotation matrix parametrized by $\theta$ and $\Lambda$ is a scaling matrix parametrized by $\rho$. For a constant \(D\) the anisotropic SPDE becomes  

\begin{equation}
\bigl(\kappa^{2}-\nabla\cdot D\nabla\bigr)\,f(\mathbf{s})=\mathcal{W}(\mathbf{s}).
\label{eq:aniso_spde}
\end{equation}

Using unit grid spacing and second‑order central differences for the
second–derivative terms,

\begin{multline}
\nabla\cdot D\nabla f_{i,j}\;\approx\;
D_{1,1}(f_{i+1,j}-2f_{i,j}+f_{i-1,j})
+D_{2,2}(f_{i,j+1}-2f_{i,j}+f_{i,j-1}) \\
+\frac{D_{1,2}}{2}\!\bigl[f_{i+1,j+1}-f_{i+1,j-1}-f_{i-1,j+1}+f_{i-1,j-1}\bigr].
\end{multline}

Substituting the above expression into \eqref{eq:aniso_spde} results in

\begin{multline}
\Bigl(\kappa^{2}+2D_{1,1}+2D_{2,2}\Bigr)f_{i,j}
\;-\;D_{1,1}(f_{i+1,j}+f_{i-1,j})
\;-\;D_{2,2}(f_{i,j+1}+f_{i,j-1})\\
\;-\;\frac{D_{1,2}}{2}\bigl[f_{i+1,j+1}-f_{i+1,j-1}-f_{i-1,j+1}+f_{i-1,j-1}\bigr]
\;=\;e_{i,j}.
\label{eq:aniso_discrete}
\end{multline}
Equation \eqref{eq:aniso_discrete} corresponds to the $3\times3$ stencil

\begin{equation}
    \begin{array}{c|c|c}
        \dfrac{D_{1,2}}{2} & -\,D_{2,2} & -\,\dfrac{D_{1,2}}{2} \\
        \hline
        -\,D_{1,1} & \kappa^{2}+2D_{1,1}+2D_{2,2} & -\,D_{1,1} \\
        \hline
        -\,\dfrac{D_{1,2}}{2} & -\,D_{2,2} & \dfrac{D_{1,2}}{2}.
    \end{array}
\label{eq:aniso_stencil}
\end{equation}

One incorporates non-stationarity by allowing $D$ and $\kappa^2$ to vary in space, as shown in Equation \eqref{eq:spde_aniso}.

\section{NON-STATIONARY DATA GENERATION DETAILS}
\label{sec:data_generation_appendix}

The synthetic data generation pipeline for the I2I networks constructs parameter fields for $\kappa^2(\mathbf{s})$, $\rho(\mathbf{s})$, and $\theta(\mathbf{s})$. The parameters have the following prior distributions: 
\begin{equation}
    \kappa^2 \sim 0.6\log\mathcal{U}(10^{-4}, 2) + 0.4\mathcal{U}(10^{-4},2),\quad \rho \sim \mathcal{U}(1, 7),\quad \theta \sim \mathcal{U}(-\frac{\pi}{2}, \frac{\pi}{2})
\end{equation}

Each parameter field is independently created by selecting one of several spatial patterns, sampling associated hyperparameters from prior distributions, sampling parameter values for the maxima and minima of the field, and optionally stacking two patterns to increase complexity. 

\subsection{Spatial Patterns}
We design eight simple yet expressive spatial patterns $p(\mathbf{s})$, intended as caricatures of realistic geophysical variability. Here we express $p$ as a generic value for $\kappa^2$, $\rho$, or $\theta$, our pattern's associated hyperparameters as $\boldsymbol{\Omega}$, and the $x$ and $y$ coordinates on the grid as $\mathbf{s}_x,\mathbf{s}_y$, respectively.

\textbf{Constant:} Uniform value across the domain.
\begin{itemize}
    \item Functional Form: $p(\mathbf{s}) = p_{\text{constant}}$, where $p_{\text{constant}}$ is a constant sampled from the prior.
\end{itemize}

\textbf{Coastline:} A sharp sigmoidal boundary, perturbed by sinusoidal "bumps".
\begin{itemize}
    \item Functional Form: 
    \begin{equation}
    p(\mathbf{s}) = p_{\text{low}} + \frac{p_{\text{high}} - p_{\text{low}}}{1 + \exp\left( -\gamma (\mathbf{s}_y - v(\mathbf{s}_x)) \right)}
    \end{equation}
    where $v(\mathbf{s}_x) = \alpha \mathbf{s}_x + \beta \sin(2\pi \omega \mathbf{s}_x) + \epsilon$, and $p_{\text{low}},p_{\text{high}}$ are the lower and higher of the two independently sampled parameter values, dictating the maxima and minima of the pattern. 
    \item Hyperparameters: $\boldsymbol{\Omega} = (\alpha, \beta, \omega, \gamma)$, sampled from $\alpha \sim \mathcal{U}(-2, 2), \beta \sim \mathcal{U}(0.1, 0.5), \omega \sim \mathcal{U}(0.4, 3), \text{and } \gamma \sim \mathcal{U}(3, 50)$.
\end{itemize}

\textbf{Taper:} Smooth transition between two values, based on a Gaussian CDF.
\begin{itemize}
    \item Functional Form: 
    \begin{equation}
    p(\mathbf{s}) = p_{\text{low}} \Psi(\mathbf{s}_x + \mathbf{s}_y; 0, \sigma) + p_{\text{high}} (1 - \Psi(\mathbf{s}_x + \mathbf{s}_y; 0, \sigma))
    \end{equation}
    where $\Psi(\cdot)$ denotes the standard normal cumulative distribution function (CDF).
    \item Hyperparameters: $\Omega = \sigma$, with $\sigma \sim \mathcal{U}(0.05, 1)$ controlling the sharpness of the transition.
\end{itemize}

\textbf{Bump:} A single, smooth, Gaussian peak.
\begin{itemize}
    \item Functional Form: 
    \begin{equation}
    p(\mathbf{s}) = p_{\text{constant}} + a_1 \exp\left( -\frac{\mathbf{s}_x^2 + \mathbf{s}_y^2}{\lambda_1} \right)
    \end{equation}
    \item Hyperparameters: $\boldsymbol{\Omega} = (a_1, \lambda_1)$, with $a_1$ controlling the peak height and $\lambda_1$ controlling spread. In all cases $\lambda_1 \sim \mathcal{U}(0.2, 0.5)$, while $a_1$ varies. When making a $\kappa^2(\mathbf{s})$ field, $a_1 \sim \mathcal{U}(0.1, 0.5)$. For $\rho(\mathbf{s})$, $a_1 \sim \mathcal{U}(0.1, 1.5)$, and for $\theta(\mathbf{s})$, $a_1 \sim \mathcal{U}(0.1, \frac{\pi}{4})$.
\end{itemize}

\textbf{Sinwave:} Periodic variation along one spatial axis.
\begin{itemize}
    \item Functional Form: 
    \begin{equation}
    p(\mathbf{s}) = 
    \begin{cases}
        p_{\text{constant}} + a \sin(\pi \omega \mathbf{s}_x), & \text{(horizontal)}\\
        p_{\text{constant}} + a \cos(\pi \omega \mathbf{s}_y), & \text{(vertical)}
    \end{cases}
    \end{equation}
    \item Hyperparameters: $\boldsymbol{\Omega} = (a, \omega, \text{orientation})$, where $\omega \sim \mathcal{U}(1.5,5)$, orientation is sampled uniformly between horizontal and vertical, and $a$ is constrained to stay within the bounds of the parameter one is constructing a field for.
\end{itemize}

\textbf{Double Bump:} Superposition of two independent Gaussian peaks located at different positions.
\begin{itemize}
    \item Functional Form: 
    \begin{align}
    p(\mathbf{s}) &= p_{\text{constant}} + a_1 \exp\left( -\frac{(\mathbf{s}_x - x_1)^2 + (\mathbf{s}_y - y_1)^2}{\lambda_1} \right) \nonumber \\
    &\quad + a_2 \exp\left( -\frac{(\mathbf{s}_x - x_2)^2 + (\mathbf{s}_y - y_2)^2}{\lambda_2} \right)
    \end{align} 
    \item Hyperparameters: $\boldsymbol{\Omega} = (a_1, a_2, \lambda_1, \lambda_2, x_1, y_1, x_2, y_2)$, with amplitudes, widths, and locations sampled as previously described in the above ``Bump'' configuration, and with centers $(x_1,y_1), (x_2,y_2)$ being randomly sampled locations within the spatial domain. 
\end{itemize}

\textbf{Double Coastline:} Weighted superposition of two independent coastline patterns.
\begin{itemize}
    \item Functional Form: 
    \begin{equation}
    p(\mathbf{s}) = p_{\text{low}} + w_1\text{Coastline}_{1}(\mathbf{s}) + w_2\text{Coastline}_{2}(\mathbf{s})
    \end{equation}
    \item Hyperparameters: $w_1 \sim \mathcal{U}(0.1,0.9), w_2 \sim \mathcal{U}(0.1, 1 - w_1)$, and each coastline has its own, independent $(\alpha, \beta, \omega, \gamma)$ parameters as described in the above ``Coastline'' configuration.
\end{itemize}

\textbf{GP-Based:} A smooth, stationary random field generated by a low-rank Gaussian process, either min-max rescaled or perturbed around a constant.
\begin{itemize}
    \item Functional Form: A realization from a Gaussian process, constructed via low-rank basis function approximation:
    \begin{equation}
    p(\mathbf{s}) = 
    \begin{cases}
        p_{\text{min}},p_{\text{max} } \text{ rescaled field}, & \text{(min-max scaling)}\\
        p_{\text{constant}} \times (1 \pm g(\mathbf{s})), & \text{(perturbation scaling)}
    \end{cases}
    \end{equation}
    where $g(\mathbf{s})$ is a normalized low-rank GP realization, $p_{\text{constant}}$ is a parameter value sampled from the prior distribution, and $p_{\text{min}},p_{\text{max} }$ represent the minimum and maximum of a parameter's prior distribution. 
    
    \item Hyperparameters: $\boldsymbol{\Omega} = (n_{\text{basis}}, \text{scaling choice})$, where $n_{\text{basis}} \sim \mathcal{U}\{6,7,\dots,32\}$ controls the number of basis functions used to generate the realization of the GP. The scaling choice is selected randomly: either
        \begin{itemize}
            \item Min-max scaling: $p(\mathbf{s})$ is rescaled to span the full prior range, or
            \item Perturbation scaling: $p(\mathbf{s})$ is a small multiplicative perturbation around $p_{\text{constant}}$, with perturbation magnitude drawn from a prior distribution. 
        \end{itemize}
    \end{itemize}

\subsection{Pattern Stacking}
To further increase the complexity of our fields, for each parameter field we randomly decide whether to linearly combine two independently generated patterns. Given fields $p_1(\mathbf{s})$ and $p_2(\mathbf{s})$, the final field is
\begin{equation}
p(\mathbf{s}) = w p_1(\mathbf{s}) + (1-w) p_2(\mathbf{s}), \quad w \sim \mathcal{U}(0.1, 0.9)
\end{equation}
Stacking introduces complex non-stationarity patterns beyond those generated by a single functional form, increasing the range of spatial patterns seen during training. 

\subsection{Resulting Synthetic Data}

\begin{figure}[h]
    \centering
    \includegraphics[width=1\textwidth]{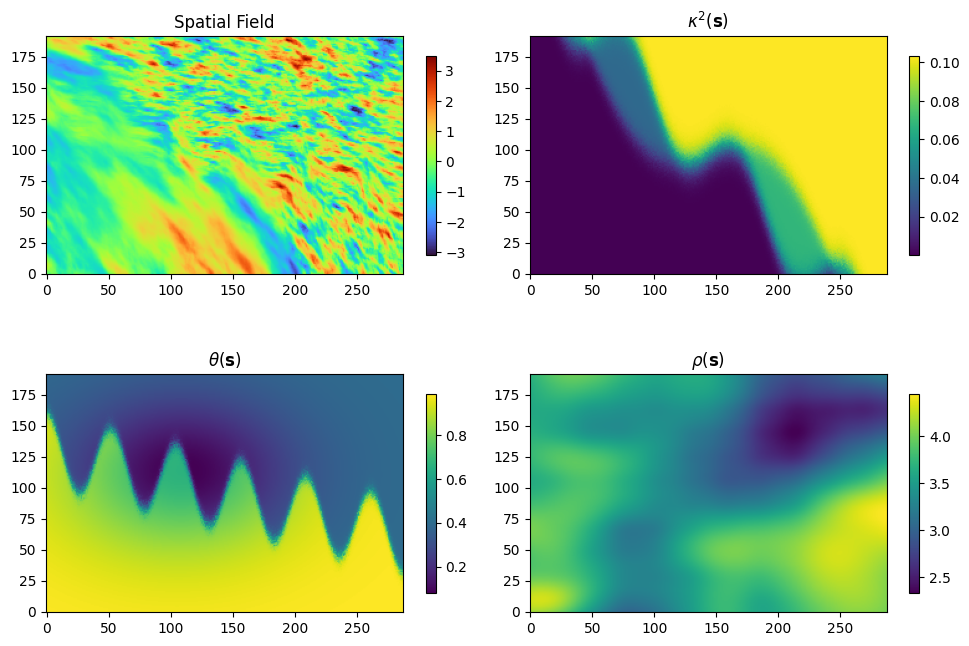}
    \caption{ The first replicate in a training sample (\textbf{top-left}), and the accompanying parameter fields that were used to generate it (\textbf{remaining}). In this instance, a Double Coastline pattern was used for $\kappa^2(\mathbf{s})$, $\theta(\mathbf{s})$ is the result of stacked Coastline and Double Bump patterns, and $\rho(\mathbf{s})$ is created with a GP-Based pattern.}
    \label{fig:appendix_plot_example_field}
\end{figure}

We repeat the process of selecting a pattern, sampling the pattern hyperparameters $\boldsymbol{\Omega}$, sampling the parameter values, and generating the resulting parameter field for $\kappa^2(\mathbf{s}), \rho(\mathbf{s}),$ and $\theta(\mathbf{s})$. We then encode all parameter fields into the SAR matrix $B$, draw white noise $\mathbf{e}\sim \mathcal{N}(0,I)$ and solve $B\,\mathbf{y} = \mathbf{e}$. We repeat this for $M$ independent draws of the white noise, resulting in a small ensemble $Y = \{\mathbf{y}^{(m)}\}^{M}_{m=1}$, with each field having the same covariance structure. We process these synthetic fields identically to the way we process the input ESM fields: we perform pixelwise standardization to ensure each pixel has a mean of 0 and a standard deviation of 1. The non-stationarity in our problem does not allow for more sophisticated normalization techniques \parencite{LKNormalizing}, which either become computationally intractable, require a stationary structure, or are not applicable in this setting. An example spatial field and its associated parameter fields can be seen in Figure \ref{fig:appendix_plot_example_field}. 

\section{FURTHER IMPLEMENTATION DETAILS}
\label{sec:implementation_appendix}

\subsection{Data and Storage}
All data generation is done in the {\tt R} programming language on a laptop with an Intel(R) Core(TM) i9-14900HX processor at 2.20 GHz, and 32GB of RAM. Fields are generated using the {\tt LatticeKrig} package, and data is compressed and stored using the {\tt hdf5} file format so it may be easily accessed in both {\tt Python} and {\tt R}. Storage proves to be more of a bottleneck as compared to data generation time: the I2I dataset (8{,}000 samples with 30 replicates) required 8 hours and occupies 108 GB, whereas the CNN dataset (80{,}000 samples with 30 replicates) is only 11 GB, and is generated in half an hour.

\subsection{Training}
All networks are implemented in PyTorch \parencite{paszke2019pytorch} and trained on a single NVIDIA RTX A6000 GPU. The networks are trained using the AdamW optimizer \parencite{adamw} for 200 epochs with a step-wise learning rate decay and early stopping after 10 epochs without validation improvement. We use mean squared error (MSE) loss, computed on normalized parameter values within the training loop. This avoids loss imbalance caused by parameter scale differences while requiring less pre or post-processing from the user.

\section{FURTHER EXPERIMENTS WITH SIMULATED DATA}

In this section, Table \ref{tab:pos_embed_metrics} contains the results of experimenting with different positional embeddings for STUN and ViT. Tables \ref{tab:appendix_i2i} and \ref{tab:appendix_cnn} contain the results of experimenting across a different number of replicates $M = \{1,5,15,30\}$ for all networks. In all tables, root mean squared error (RMSE), mean absolute error (MAE), and normalized RMSE (NRMSE) are calculated, along with the structural similarity index measure (SSIM) and peak signal to noise ratio (PSNR) in decibels (db) image comparison metrics. 

\begin{table}[]
\caption{Results on simulated test data across different types of positional embeddings for both STUN and ViT. Best results for each architecture are in bold.}
\label{tab:pos_embed_metrics}
\centering
\begin{tabular}{lllccccc}
\toprule
\multirow{2}{*}{Net}&\multirow{2}{*}{Embedding}&\multirow{2}{*}{Param}&\multicolumn{5}{c}{Metrics}\\
\cmidrule(lr){4-8}
&&&RMSE $\downarrow$&MAE $\downarrow$&SSIM $(\uparrow \text{to } 1)$& PSNR $\uparrow$&NRMSE $\downarrow$\\
\midrule
\multirow{12}{*}{ViT}
	&\multirow{3}{*}{None}&$\kappa^2$&0.398&0.279&0.252&10.9&0.040\\
	&&$\rho$      &0.711&0.525&0.277&6.41&0.118\\
	&&$\theta$    &0.249&0.124&0.251&13.6&0.083\\
\cmidrule(lr){2-8}
	&\multirow{3}{*}{Sinusoidal}&$\kappa^2$&0.393&0.274&0.275&10.8&0.040\\
	&&$\rho$      &0.913&0.723&0.294&3.61&0.152\\
	&&$\theta$    &0.280&0.143&0.253&12.4&0.094\\
\cmidrule(lr){2-8}
	&\multirow{3}{*}{Learned}&$\kappa^2$&\textbf{0.371}&\textbf{0.243}&0.291&\textbf{12.1}&\textbf{0.037}\\
	&&$\rho$      &0.920&0.715&0.288&3.30&0.153\\
	&&$\theta$    &0.351&0.193&0.235&9.83&0.117\\
\cmidrule(lr){2-8}
	&\multirow{3}{*}{Rotary}&$\kappa^2$&0.374&0.250&\textbf{0.310}&11.9&0.038\\
	&&$\rho$      &\textbf{0.625}&\textbf{0.465}&\textbf{0.337}&\textbf{7.01}&\textbf{0.104}\\
	&&$\theta$    &\textbf{0.204}&\textbf{0.103}&\textbf{0.328}&\textbf{15.1}&\textbf{0.068}\\
\midrule
\multirow{12}{*}{STUN}
	&\multirow{3}{*}{None}&$\kappa^2$&0.198&0.118&\textbf{0.489}&18.4&0.020\\
	&&$\rho$      &0.311&0.213&\textbf{0.548}&14.2&0.052\\
	&&$\theta$    &0.096&0.046&\textbf{0.604}&21.7&0.032\\
\cmidrule(lr){2-8}
	&\multirow{3}{*}{Sinusoidal}&$\kappa^2$&0.190&\textbf{0.113}&0.483&\textbf{18.7}&0.019\\
	&&$\rho$      &0.302&\textbf{0.202}&0.542&\textbf{14.7}&0.050\\
	&&$\theta$    &0.126&0.050&0.592&21.5&0.042\\
\cmidrule(lr){2-8}
	&\multirow{3}{*}{Learned}&$\kappa^2$&0.204&0.122&0.462&18.0&0.021\\
	&&$\rho$      &0.309&0.217&0.522&13.9&0.052\\
	&&$\theta$    &\textbf{0.090}&0.045&0.595&21.8&\textbf{0.030}\\
\cmidrule(lr){2-8}
	&\multirow{3}{*}{Rotary}&$\kappa^2$&\textbf{0.189}&0.117&0.469&18.4&\textbf{0.019}\\
	&&$\rho$      &\textbf{0.302}&0.212&0.531&14.3&\textbf{0.050}\\
	&&$\theta$    &0.091&\textbf{0.044}&0.594&\textbf{21.9}&0.030\\
\bottomrule
\end{tabular}
\end{table}

\begin{table}[]
\caption{Results on simulated test data across varying replicates for image-to-image networks.}
\label{tab:appendix_i2i}
\centering
\begin{tabular}{lllccccc}
\toprule
\multirow{2}{*}{Net} & \multirow{2}{*}{Reps} & \multirow{2}{*}{Param}
  & \multicolumn{5}{c}{Metrics} \\
\cmidrule(lr){4-8}
 & &  & RMSE $\downarrow$ & MAE $\downarrow$ & SSIM $(\uparrow \text{to } 1)$ & PSNR $\uparrow$ & NRMSE $\downarrow$ \\
\midrule
\multirow{12}{*}{UNet}
	& \multirow{3}{*}{1}  & $\kappa^2$ & 0.195& 0.125& 0.447& 17.6& 0.020\\
	&                    & $\rho$      & 0.354& 0.250& 0.534& 13.1& 0.059\\
	&                    & $\theta$    & 0.111& 0.052& 0.571& 20.8& 0.037\\
\cmidrule(lr){2-8}
	& \multirow{3}{*}{5}  & $\kappa^2$ & 0.199& 0.125& 0.438& 17.7& 0.020\\
	&                    & $\rho$      & 0.319& 0.223& 0.527& 13.8& 0.053\\
	&                    & $\theta$    & 0.115& 0.053& 0.546& 20.5& 0.038\\
\cmidrule(lr){2-8}
	& \multirow{3}{*}{15} & $\kappa^2$ & 0.180& 0.110& 0.503& 18.9& 0.018\\
	&                    & $\rho$      & 0.294& 0.207& 0.592& 14.6& 0.049\\
	&                    & $\theta$    & 0.102& 0.044& 0.628& 22.3& 0.034\\
\cmidrule(lr){2-8}
	& \multirow{3}{*}{30} & $\kappa^2$ & 0.201& 0.124& 0.447& 17.9& 0.020\\
	&                    & $\rho$      & 0.308& 0.214& 0.517& 14.1& 0.051\\
	&                    & $\theta$    & 0.087& 0.046& 0.582& 21.3& 0.029\\
\midrule
\multirow{12}{*}{ViT}
	& \multirow{3}{*}{1}  & $\kappa^2$ & 0.471& 0.303& 0.296& 10.5& 0.048\\
	&                    & $\rho$      & 0.771& 0.588& 0.328& 5.36& 0.129\\
	&                    & $\theta$    & 0.237& 0.113& 0.323& 14.3& 0.079\\
\cmidrule(lr){2-8}
	& \multirow{3}{*}{5}  & $\kappa^2$ & 0.377& 0.243& 0.326& 12.3& 0.038\\
	&                    & $\rho$      & 0.633& 0.470& 0.358& 7.06& 0.106\\
	&                    & $\theta$    & 0.211& 0.098& 0.359& 15.6& 0.071\\
\cmidrule(lr){2-8}
	& \multirow{3}{*}{15} & $\kappa^2$ & 0.354& 0.229& 0.324& 12.6& 0.036\\
	&                    & $\rho$      & 0.594& 0.425& 0.350& 7.73& 0.099\\
	&                    & $\theta$    & 0.206& 0.099& 0.340& 15.4& 0.069\\
\cmidrule(lr){2-8}
	& \multirow{3}{*}{30} &$\kappa^2$&0.374&0.250&0.310&11.9&0.038\\
	&&$\rho$      &0.625&0.465&0.337&7.01&0.104\\
	&&$\theta$    &0.204&0.103&0.328&15.1&0.068\\
\midrule
\multirow{12}{*}{STUN}
	& \multirow{3}{*}{1}  & $\kappa^2$ & 0.189& 0.124& 0.470& 17.7& 0.019\\
	&                    & $\rho$      & 0.351& 0.243& 0.535& 13.3& 0.058\\
	&                    & $\theta$    & 0.097& 0.047& 0.593& 21.3& 0.033\\
\cmidrule(lr){2-8}
	& \multirow{3}{*}{5}  & $\kappa^2$ & 0.180& 0.109& 0.502& 18.8& 0.018\\
	&                    & $\rho$      & 0.290& 0.198& 0.554& 14.8& 0.048\\
	&                    & $\theta$    & 0.100& 0.044& 0.632& 22.2& 0.033\\
\cmidrule(lr){2-8}
	& \multirow{3}{*}{15} & $\kappa^2$ & 0.188& 0.114& 0.483& 18.6& 0.019\\
	&                    & $\rho$      & 0.303& 0.209& 0.551& 14.4& 0.051\\
	&                    & $\theta$    & 0.102& 0.047& 0.589& 21.6& 0.034\\
\cmidrule(lr){2-8}
	& \multirow{3}{*}{30} & $\kappa^2$ & 0.189 &0.117&0.469&18.4&0.019 \\
	&                    & $\rho$      &0.302&0.212&0.531&14.3&0.050 \\
	&                    & $\theta$    &0.091&0.044&0.594&21.9&0.030\\
\bottomrule
\end{tabular}
\end{table}

\begin{table}[]
\caption{Results on simulated test data across varying replicates for local CNNs.}
\label{tab:appendix_cnn}
\centering
\begin{tabular}{lllccccc}
\toprule
\multirow{2}{*}{Net} & \multirow{2}{*}{Reps} & \multirow{2}{*}{Param}
  & \multicolumn{5}{c}{Metrics} \\
\cmidrule(lr){4-8}
 & &  & RMSE $\downarrow$ & MAE $\downarrow$ & SSIM $(\uparrow \text{to } 1)$ & PSNR $\uparrow$ & NRMSE $\downarrow$ \\
\midrule
\multirow{12}{*}{CNN9}
 & \multirow{3}{*}{1}  & $\kappa^2$ & 1.01& 0.637& 0.069& 4.64& 0.101\\
 &                    & $\rho$      & 1.39& 1.09& 0.02& -0.980& 0.231\\
 &                    & $\theta$    & 0.535& 0.295& 0.072& 5.27& 0.179\\
 \cmidrule(lr){2-8}
 & \multirow{3}{*}{5}  & $\kappa^2$ & 0.983& 0.547& 0.129& 6.97& 0.099\\
 &                    & $\rho$      & 1.03& 0.760& 0.052& 2.02& 0.172\\
 &                    & $\theta$    & 0.346& 0.149& 0.217& 10.5& 0.116\\
 \cmidrule(lr){2-8}
 & \multirow{3}{*}{15} & $\kappa^2$ & 0.949& 0.512& 0.160& 7.81& 0.096\\
 &                    & $\rho$      & 0.962& 0.689& 0.070& 2.90& 0.160\\
 &                    & $\theta$    & 0.330& 0.135& 0.300& 11.3& 0.110\\
 \cmidrule(lr){2-8}
 & \multirow{3}{*}{30} & $\kappa^2$ & 0.963& 0.508& 0.172& 8.09& 0.097\\
 &                    & $\rho$      & 0.937& 0.660& 0.085& 3.28& 0.156\\
 &                    & $\theta$    & 0.316& 0.121& 0.347& 11.7& 0.106\\
 \midrule
 \multirow{12}{*}{CNN17}
 & \multirow{3}{*}{1}  & $\kappa^2$ & 0.766& 0.480& 0.086& 6.45& 0.077\\
 &                    & $\rho$      & 1.27& 0.960& 0.026& -0.062& 0.212\\
 &                    & $\theta$    & 0.443& 0.200& 0.147& 7.60& 0.148\\
 \cmidrule(lr){2-8}
 & \multirow{3}{*}{5}  & $\kappa^2$ & 0.828& 0.443& 0.169& 8.52& 0.084\\
 &                    & $\rho$      & 1.03& 0.714& 0.081& 2.75& 0.171\\
 &                    & $\theta$    & 0.317& 0.126& 0.289& 11.1& 0.106\\
 \cmidrule(lr){2-8}
 & \multirow{3}{*}{15} & $\kappa^2$ & 0.730& 0.377& 0.222& 9.94& 0.074\\
 &                    & $\rho$      & 0.939& 0.614& 0.123& 4.00& 0.157\\
 &                    & $\theta$    & 0.266& 0.097& 0.411& 12.0& 0.089\\
 \cmidrule(lr){2-8}
 & \multirow{3}{*}{30} & $\kappa^2$ & 0.764& 0.397& 0.212& 9.63& 0.077\\
 &                    & $\rho$      & 0.994& 0.669& 0.103 & 3.37& 0.166\\
 &                    & $\theta$    & 0.293& 0.116& 0.349& 11.5& 0.098\\
 \midrule
 \multirow{12}{*}{CNN25}
 & \multirow{3}{*}{1}  & $\kappa^2$ & 0.806& 0.488& 0.111& 6.81& 0.081\\
 &                    & $\rho$      & 1.28& 0.949& 0.036& 0.164& 0.214\\
 &                    & $\theta$    & 0.418& 0.179& 0.215& 8.21& 0.140\\
 \cmidrule(lr){2-8}
 & \multirow{3}{*}{5}  & $\kappa^2$ & 0.767& 0.409& 0.166& 9.0& 0.077\\
 &                    & $\rho$      & 1.09& 0.736& 0.088& 2.77& 0.181\\
 &                    & $\theta$    & 0.340& 0.124& 0.316& 10.2& 0.114\\
 \cmidrule(lr){2-8}
 & \multirow{3}{*}{15} & $\kappa^2$ & 0.732& 0.372& 0.249& 10.2 & 0.074\\
 &                    & $\rho$      & 1.03& 0.669& 0.150& 3.71& 0.172\\
 &                    & $\theta$    & 0.279& 0.107& 0.445& 11.5& 0.093\\
 \cmidrule(lr){2-8}
 & \multirow{3}{*}{30} & $\kappa^2$ & 0.743 & 0.378 & 0.256 & 10.1 & 0.075\\
 &                    & $\rho$      & 1.03& 0.676& 0.144& 3.55& 0.172\\
 &                    & $\theta$    & 0.272& 0.106& 0.434& 11.8& 0.091\\
\bottomrule
\end{tabular}
\end{table}

\newpage 

\section{FURTHER CLIMATE APPLICATION RESULTS}
\label{appendix:clim}

We use data from three climate models: the Max Planck Institute Earth System Model at Low Resolution (MPI-ESM 1.2LR), the Community Earth System Model (CESM1), and the European Community Earth System Model (EC-Earth3). We pass the standardized ensembles of temperature sensitivity anomalies into each of our estimation networks: STUN, UNet, ViT, CNN25, CNN17, and CNN9. Using the predicted parameters, we generate 1{,}000 synthetic fields from each model. Generation is performed using the LatticeKrig R package \parencite{LKrigpackage}, which accommodates cylindrical geometry and ensures the correct treatment of spatial distances under the Mercator projection. 

To evaluate how well each synthetic ensemble preserves the spatial correlation structure, we randomly select 50 anchor locations and compute the corresponding 50 rows from each simulated ensemble’s empirical correlation matrix. These are then compared to the 50 rows from the original ensemble’s correlation matrix using root mean squared error (RMSE). We systematically pair each I2I network with a local neural estimator of comparable rank (e.g., STUN vs. CNN25, UNet vs. CNN17), and compare their RMSEs across the same anchor locations. As the anchor locations are fixed for all pairings, we conduct paired t-tests between matched RMSEs to compare performance. 

Our null hypothesis for each paired comparison is that the mean RMSE difference equals zero, \(H_0\!:\,\mu_d = 0\), where \(d = \text{RMSE}_{\text{I2I}} - \text{RMSE}_{\text{CNN}}\). The one‑sided alternative, \(H_1\!:\,\mu_d < 0\), states that the I2I emulator attains a lower RMSE than its CNN counterpart.
All tests are conducted with a significance level \(\alpha = 0.01\). Results are summarized in Table~\ref{tab:t_tests_clim}. For eight of the nine comparisons, the null hypothesis is rejected and the $99\%$ confidence interval is entirely negative, demonstrating that I2I networks tend to produce significantly more accurate correlation structure estimates than their corresponding local CNNs.

\begin{table}[H]
  \caption{Paired t-test results comparing I2I-based emulators to corresponding CNN-based emulators on average RMSE across the same 50 anchor locations. CI denotes the 99\% confidence interval of the paired differences.}
  \label{tab:t_tests_clim}
  \centering
  \begin{tabular}{llccccc}
    \toprule
    \multirow{2}{*}{Climate model} & \multirow{2}{*}{Network Pair} &
    \multicolumn{5}{c}{Paired t-test metrics} \\
    \cmidrule(lr){3-7}
     & & $\mu_{\text{I2I}}$ $\downarrow$ & $\mu_{\text{local}}$ $\downarrow$ & $\mu_d$ & 99\% CI & $p_{\text{value}}$ \\
    \midrule
    \multirow{3}{*}{MPI-ESM (192 $\times$ 96)}
        & STUN vs. CNN25 & 0.202 & 0.210 & $-0.008$ & ($-\infty$, $-0.004$] & $2.2\times10^{-6}$ \\
        & UNet vs. CNN17 & 0.203 & 0.214 & $-0.011$ & ($-\infty$, $-0.005$] & $1.0\times10^{-5}$ \\
        & ViT vs. CNN9   & 0.208 & 0.247 & $-0.039$ & ($-\infty$, $-0.028$] & $7.1\times10^{-12}$ \\
    \cmidrule(lr){1-7}
    \multirow{3}{*}{CESM1 (288 $\times$ 192)}
        & STUN vs. CNN25 & 0.229 & 0.243 & $-0.013$ & ($-\infty$, $-0.007$] & $1.1\times10^{-6}$ \\
        & UNet vs. CNN17 & 0.230 & 0.239 & $-0.008$ & ($-\infty$, $-0.004$] & $3.4\times10^{-5}$ \\
        & ViT vs. CNN9   & 0.228 & 0.244 & $-0.016$ & ($-\infty$, $-0.008$] & $5.9\times10^{-6}$ \\
    \cmidrule(lr){1-7}
    \multirow{3}{*}{EC-Earth3 (512 $\times$ 256)}
        & STUN vs. CNN25 & 0.271 & 0.285 & $-0.014$ & ($-\infty$, $-0.005$] & $1.6\times10^{-4}$ \\
        & UNet vs. CNN17 & 0.275 & 0.288 & $-0.012$ & ($-\infty$, $-0.004$] & $2.7\times10^{-4}$ \\
        & ViT vs. CNN9   & 0.283 & 0.304 & $-0.021$ & ($-\infty$, $0.0003$] & $1.1\times10^{-2}$ \\
    \bottomrule
  \end{tabular}
\end{table}

\end{document}